%% file: paper.tex
\documentclass{ecai}
\usepackage{times}
\usepackage[hyphens]{url}
\usepackage{graphicx}
\usepackage{algorithm}
\usepackage[noend]{algpseudocode}
\usepackage{bm}
\usepackage{array}
\usepackage{amsfonts}

\input{pm_macros}

\usepackage{subfig}
\usepackage{float}
\usepackage{bbm}
\usepackage{amsmath}
\usepackage{lipsum}  
\usepackage{amssymb}
\usepackage{siunitx}
\usepackage{mathtools}
\usepackage{multirow}
\DeclarePairedDelimiter{\ceil}{\lceil}{\rceil}

\usepackage{color}

\def \calK{{\cal K}}
\def \calM{{\cal M}}
\def \calY{{\cal Y}}
\def\argmax{\operatornamewithlimits{argmax}}
\def \random{{\tt random}\xspace}
\def \devel{{\tt devel}\xspace}
\def \dev{{\tt devel}\xspace}
\def \fullmethod{\textsc{FULLower}\xspace}
\def \method{\textsc{Follower}\xspace}
\def \fullmethods{\textsc{FULLower}s\xspace}
\def \methods{\textsc{Follower}s\xspace}
\def \testset{S_{T_e}}
\def \same{\texttt{same}\xspace}

\title{Continual egocentric object recognition}
\author{
    Luca Erculiani \institute{University of Trento, Italy, email: {\it name.surname}@unitn.it}
    \and 
    Fausto Giunchiglia {\normalfont\textsuperscript{1}}\institute{College of Computer Science and Technology, Jilin University, Changchun, China, CN}
    \and
    Andrea Passerini {\normalfont\textsuperscript{1}}
}

\begin{document}

\maketitle

\begin{abstract}

  We present a framework capable of  tackilng 
  the problem of continual object recognition in
  a setting which resembles that under which humans see and
  learn. 
  This setting has a set of unique characteristics: it assumes an
  egocentric point-of-view bound to the needs of a single person, which implies a
  relatively low diversity of data and a cold start with no data; it
  requires to operate in an open world, where new objects can be
  encountered {\em at any time}; supervision is scarce and has to be
  solicited to the user, and completely {\em unsupervised recognition}
  of new objects should be possible.   
  Note that this setting differs
  from the one addressed in the open world recognition literature,
  where supervised feedback is always requested to be able to
  incorporate new objects. 
  We propose a first solution to this problem in
  the form of a memory-based incremental framework that 
  is capable of
  storing information of each and any object it encounters, while
  using the supervision of the user to learn to discriminate between
  known and unknown objects. Our approach is   
  based on four main features: the use of time and space persistence
  (i.e., the appearance of objects changes relatively slowly), the use
  of similarity as the main driving principle for object recognition
  and novelty detection, the progressive introduction of new objects
  in a developmental fashion and the selective elicitation of user
  feedback in an online active learning fashion.
  Experimental results
  show the feasibility of open world, generic object recognition, the
  ability to recognize, memorize and re-identify new objects even in
  complete absence of user supervision, and the utility of persistence
  and incrementality in boosting performance.
\end{abstract}

\section{Introduction}
Over the last few years Deep Neural Networks led to massive
improvements for the tasks of object detection and recognition in
images~\cite{he2016deep,redmon2016you}. The main application scenario
for this work is the use of large sets of photos, most of the time
collected from the Web, to reliably identify objects from an
increasingly large but static hierarchy of
classes~\cite{russakovsky2015imagenet}.
One of the key factors of the success of these approaches has been the
availability of large datasets of (annotated) images and videos
\cite{deng2009imagenet}.

Our motivating scenario is quite different. We are interested in
recognizing objects in a setting which resembles that under which
humans see and perceive the world.  This problem is of high relevance
in all those applications where there is a wearable camera (e.g., in
the glasses) which generates images or videos whose recognition can be
used to support the user in her local needs (e.g., everyday life,
working tasks). The main innovative characteristics in this setting
are: (i) it assumes an egocentric setting~\cite{SmiSlo2017}, where the
input is the point-of-view of a single person (i.e.,
the data has low diversity and high correlation); 
(ii) there is a continuous flow of input data, with
new objects appearing all the time (i.e., we assume the agent operates
in an open world); (iii) recognition should be able to go as deep as
instance-level re-identification (e.g. recognizing my own mug); (iv)
supervision is scarce and should be solicited to the user when needed,
also accounting for entirely autonomous identification and processing
of new objects.

This scenario contrasts with the typical setting in which deep
learning architectures shine. Incorporating novel classes is a
notoriously hard problem for deep networks. The way in which these
networks are trained drives them to learn models that implicitly
follow the closed world assumption, and trying to dynamically expand
their capabilities negatively affects previous knowledge (the
so-called catastrophic forgetting~\cite{goodfellow2014empirical}). 
While substantial progresses have been made in fields like continual
lifelong learning~\cite{parisi2018continual} and few-shot
learning~\cite{koch2015siamese}, the state-of-the-art algorithms in
this field are far from being able to match the capabilities of
humans.
We argue that a key factor for this gap is the way these algorithms
are exposed to the data during training, with respect to what happens
for humans and other animals.
Humans experience the world via a continuous stream of highly
correlated visual stimuli, initially focused on very few objects. This
enables them to progressively acquire an understanding of the
difference ways in which objects can appear, and on the similarities
and differences between objects.

On these premises, the solution proposed in this paper is based on the
following intuitions:

\begin{itemize}
\item Introduce persistence as a key element for instance-level
  labeling. By this we mean the fact that when we see an object as
  part of a video the object will change very slowly, allowing to
  identify {\em visual invariances} useful for subsequence
  recognition. This is thought to be one of the key aspects for early
  visual learning in children~\cite{foldiak91,li_unsupervised_2008}.

\item Use similarity as the main driving principle for object
  recognition and novelty detection. This is consistent with the
  recent trend in
  few-shot~\cite{koch2015siamese,snell2017prototypical} and
  open-world~\cite{bendale2015towards,rudd2018extreme}) learning, and
  we extend it here the autonomous recognition of new objects.

\item Progressively introduce novel objects in a developmental
  fashion~\cite{Bengio:2009,SmiSlo2017}, and provide supervision
  on-demand in an online active learning fashion.

\end{itemize}

The proposed algorithm, which we named \method, performs online
open-world object recognition over a continuous stream of videos of
objects, taking an egocentric point-of-view bound to a specific
user. Our experimental evaluation is promising, and highlights the
importance of persistence, especially when dealing with instance-level
recognition, the ability to identify and memorize novel objects even
in complete absence of user supervision, and the role of developmental
learning in boosting performance.

\section{Related Work}

Our framework is related to the task of continual lifelong
learning~\cite{parisi2018continual}. In continual lifelong learning
the learner is presented as sequence of study-sessions, each focused
on a novel class, and the goal of the learner is being able to
incorporate new classes without forgetting previously learned
ones. The main approaches to address the problem are constraining
model updates to limit interference with previous
knowledge~\cite{zenke2017continual,yoon2018lifelong}, and learning
separate subsystems to handle short- and long-term
memory~\cite{gepperth2016bio}. While still aiming at continual
learning, our framework is substantially different as there is no
separate teaching session for each class, but rather a continuous
stream of objects with on-demand supervision, the focus is on
recognizing individual objects, 
and we assume a regime of scarce labelled data.

Few-shot learning is the problem of learning a task when very few data
is available for training.  Existing research either focus on
combining metric and class prototype
learning~\cite{koch2015siamese,triantafillou2017few}, or frame the
problem as meta-learning~\cite{santoro2016meta,mishra2018simple}.  As
for continual lifelong learning, these approaches typically assume a
set of class-specific teaching sessions.  Our solution shares the idea
of using similarity-based solutions to cope with data scarcity, but
frames it in an online, active learning setting where new objects can
appear at all times, in a fully open-world scenario.

Getting rid of the closed-world assumption is a recent research trend
in the machine learning community. The problem was first framed in
terms of {\em open set} learning, i.e. learn to identify examples of unknown classes%
~\cite{scheirer2013toward,bendale2016towards}.  More
recently, {\em open world} learning has been introduced, where the
learner should both identify examples of unknown classes, and be able
to add them in the future as a novel class, if new external
supervision is made available. As happens for most few-shot learning
strategies, many existing approaches for open world learning are
similarity-based~\cite{bendale2015towards,rudd2018extreme}.
\cite{mu2017classification} proposed a method to 
partition the feature domain in subspaces, each associated with a 
local classifier and an anomaly detector, 
in order to both reduce the update effort while being able to reject unknown classes.
Still, these works assume that novel classes are incorporated via
class-specific training sessions, and their main objective is
minimizing the effort required to update their internal
representations.  Our solution adapts this similarity-based principle
to deal with user supervision in an online active learning fashion,
and to work in an instance-level object recognition scenario.

The task of discriminating objects at an instance level is usually
referred to as \emph{re-identification} in the computer vision
community. The most prominent applications in this field are related
to the identification of humans~\cite{wu2018exploit}, with some works
focusing on vehicle re-identification~\cite{shen2017learning}.  Most
of these approaches are again similarity-based, with embeddings of
images or videos learned using Siamese Neural
Networks~\cite{koch2015siamese}. The main drawbacks of these
approaches is the fact that they were specifically developed to deal
with single categories of objects (humans or vehicles), while we aim
at re-identification of generic objects. Furthermore,
re-identification assumes that the object to be re-identified is
already present in memory, while we focus on an open-world setting
where new objects are continuously presented and need to be
identified, stored and re-identified at subsequent encounters.

What we share with the re-identification literature is the idea of
using videos of objects for identification/recognition, rather than
for e.g. activity recognition or tracking. This is enables to exploit
space and time persistence to build a better picture of the
characteristics of an object
. Indeed, recent works \cite{misra2016shuffle,lee2017unsupervised}
have shown how unsupervised pre-training with videos enables to learn
useful representations for image classification tasks, and how few
videos of a given class of objects are sufficient to train a detector
for that class~\cite{prest2012learning}. These works suggest that
inside a video stream there is an incredible amount of information,
that can be used to model the recorded subjects.  We aim to leverage
on this information to build an effective few-shot object recognition
algorithm from videos.

In the robotics field, a number of works have focused on studying
approaches for human-guided interactive learning
\cite{wersing2007online,kirstein2009vision,skovcaj2011system,nakamura2012learning,oliveira20163d}.
In this setting a machine is trained to recognize objects that are
manipulated by a human \cite{wersing2007online,kirstein2009vision}, or
directly by the robot
\cite{nakamura2012learning}, asking supervision to the user if the object is considered unknown. 
While sharing similarities with our setting, these approaches are
focused on recognizing what is known and rejecting what is unknown,
and always require human intervention in order to expand of the pool
of known objects. In contrast, we aim at building systems that can
autonomously discriminate the unknown and integrate it with the
previous knowledge, expanding the pool of recognized entities even
without human intervention.

\section{The Recognition Framework}
\label{sec:framework}

We first describe the setting in which the agent in expected to
operate, and then present a learning algorithm which we consider
appropriate for the setting.

\subsection{The Setting}
\label{subsec:setting}

As explained in the introduction, we focus on open-world, incremental
instance-level object recognition, with an egocentric view bound to a specific
user. Note that we assume maximal granularity, meaning that each
object has its own class, 
and the goal of the algorithm is to cluster
input data rather than assigning a label to each sample (see
Section~\ref{sec:conclusion} for a hint on how to generalize to mixed
instance-level and class-level recognition). The information is made
available to the learning agent via a continuous stream of data. Each
sample in this stream is a short video, focused on a certain
object. The goal of the agent is, for each new video, to determine if
it contains an object it has already seen, or if it shows an object it
sees for the first time.  After the first encounter, the agent should
add the new object to the pool of objects it knowns, so as to
recognize it in the future. The agent can query the user in order to
gather supervision on instances it considers uncertain and prevent
taking wrong decisions, in an online active learning fashion.

\subsection{The Algorithm}

\begin{algorithm}[t]
    \caption{Open world egocentric object recognition\label{alg:incremental}}
	\raggedright
    \textbf{Input} : a stream of videos $S$ \\ 
	\textbf{Input} : a real value $\alpha \in [0,1]$ \\ 
    \begin{algorithmic}[1]
    \Procedure{\method}{$S,\alpha$}

    \State $\calM \gets \emptyset$; $\calK \gets \emptyset$
	\While{$s \gets \textsf{next}(S)$}
	\State $r_s \gets \textsf{embedVideo}(s)$
	    \State $r_{min}, l_{min} \gets \textsf{getNearestNeighbour}(r_s, \calM) $
	    \State $\delta \gets d(r_s,r_{min}) $
	    \State $\lambda_{l},\lambda_{u}  \gets \textsf{getDecisionThresholds}(\calK, \alpha)$	   
            \If {$\delta < \lambda_l$}
            \State $l_{s} \gets l_{min} $
            \ElsIf {$ \delta > \lambda_u  $}
            \State $l_{s} \gets \textsf{newId}()$ 
	    \Else
		\State $y \gets \textsf{askUserSupervision}(r_s,r_{min})$
		\If {$y$}
		\State $l_{s} \gets l_{min} $
		\Else
		\State $l_{s} \gets \textsf{newId}()$ 
		\EndIf

		\State $\calK \gets \calK \cup \{ \langle \delta,y \rangle\} $

	\EndIf
	\State $\calM \gets \calM \cup \{\langle r_s, l_s \rangle\} $
	\EndWhile
    \EndProcedure
    \end{algorithmic}
\end{algorithm}

\begin{figure*}
	\centering
	\includegraphics[width = 1\textwidth]{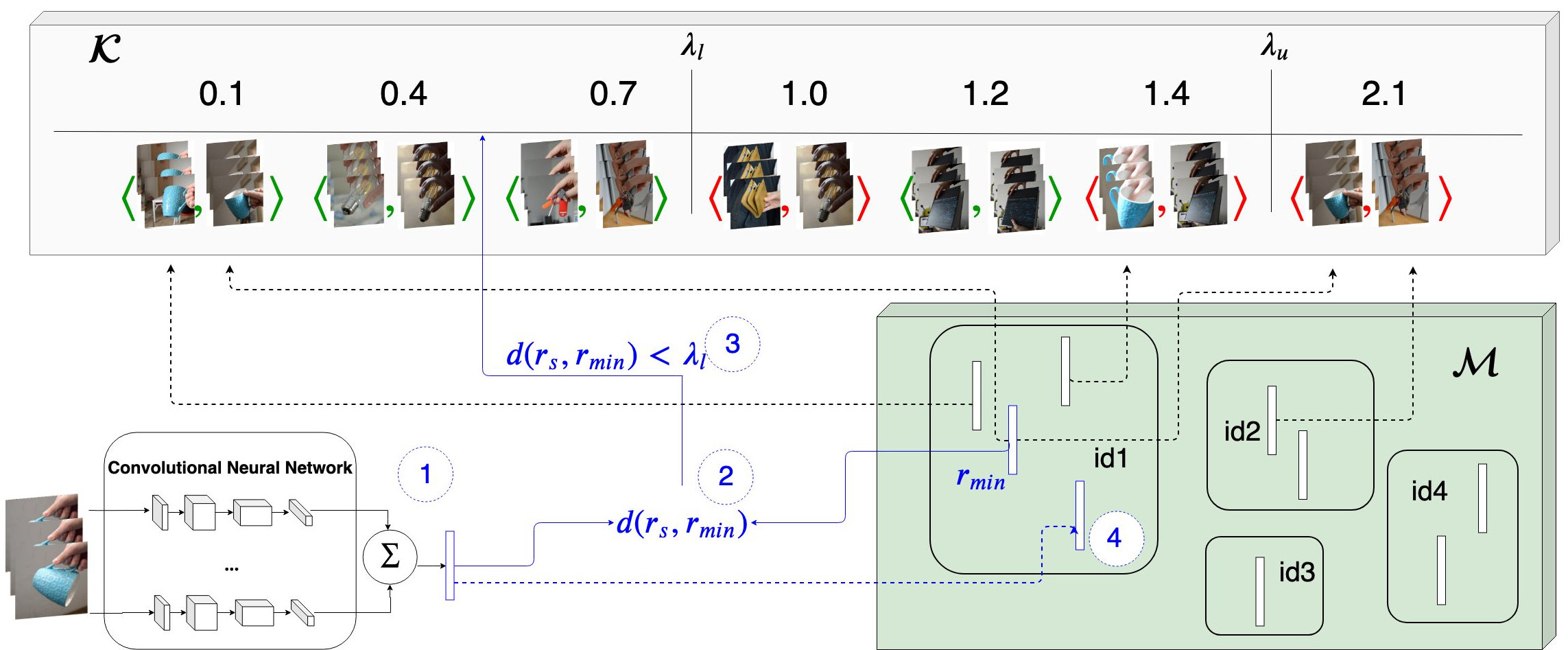}
	\label{fig:scheme}
	\caption{Graphical representation showing an iteration of our
          algorithm. In black the state of the algorithm at the end of
          the previous iteration. In green and red the tuples of same
          and different objects inside $\cal K$.  In blue the
          operations that occur at the current iteration.  The new
          video $s$ is converted in an embedding $r_s$ (1) and its
          nearest neighbor $r_{min}$ is identified (2). Given that
          their distance is lower than $\lambda_l$ (3), $r_s$ is added
          to the memory $\cal M$ with the same identifier of $r_{min}$
          (4), and the iteration finishes. No user feedback is
          required in this iteration.}
\end{figure*}

The pseudocode of the algorithm is shown in
Algorithm~\ref{alg:incremental}. It takes as input a stream of video
sequences $S$ and a single parameter $\alpha \in [0,1]$ that
represents the amount of effort the user is expected to provide on
average, in terms of number of queries. 
The $\alpha$ value, albeit requiring to be set before training, 
should not be considered an hyperparameter of the model, but rather
a value that relates with the expected effort that the user is willing
to offer while supervising the model.

The algorithm consists of an
iterative routine that receives as input a new video $s$ at each
iteration.

The assumption of persistence enables the model to assume 
each frame contains the same object, and thus to safely combine
together the information contained in different shots inside the same sequence.
The new video is transformed in a fixed size representation
$r_s \in R^n$, by generating embeddings for each frame using ResNet152
\cite{he2016deep}, and computing the mean of the embeddings of the
frames.  
In principle more advanced architectures can be used for this aggregation.
In practice the combination of constraints of the setting 
(online training, lifelong learning, low number of sequences)
makes the training of complex architectures,
like for instance recurrent networks, hard and underperforming 
with respect to the mean vector.

The resulting representation is compared
with a collection of representations of past videos, that the
algorithm has stored in its memory.  The algorithm then decides if
there exists a representation in memory that contains the very same
object, at instance level. 
It starts by computing the distance $\delta$ between the new
representation and its nearest neighbor $r_{min}$ (in memory). Based
on this distance it needs to decide among three possible options: 1)
the object is the same as the one retrieved from memory; 2) the object
is a new one; 3) there is not enough evidence to make a decision, and
the user feedback is needed.  The choice among the three options is
made by comparing the distance $\delta$ with a pair of thresholds
$\lambda_{l},\lambda_{u}$, computed using a procedure described later
on. If $\delta < \lambda_l$, the object is recognized as already seen
(option 1), and it is assigned the same identifier $l_{min}$ of its
nearest neighbour. If $\delta > \lambda_{u}$ the object is considered
a new one (option 2) and it is assigned a brand new
identifier. Otherwise, a user feedback is requested (option 3). In
this latter case, the user is asked whether $r_s$ and $r_{min}$ are
the same objects, and the identifier of the object is updated
according to the user feedback.  Finally the new object-label pair is
added to the memory 
\footnote{In order to help the user recognize previous objects and provide supervision, 
auxiliary information can be stored together with the representations. Examples 
are raw frames from the original videos, metadata to tag the time/position 
of the encounters or labels suggested by the user.}
$\calM$.  If the algorithm decided to request the user
supervision, the answer to its query is stored in another memory
$\calK$, together with the distance $\delta$.  Then the whole process
is repeated with a new video.

As described above, the whole decision process depends on the values
of the two thresholds $\lambda_{l}$ and $\lambda_{u}$. These
thresholds are estimated at each iteration by using the information
provided by the user, in the form of the collection
$\calK = \{ \langle \delta_i, y_i \rangle \mid 1 < i < |\calK| \}$ of
distances between pairs of objects $\delta_i = d(r_i, r_i')$, coupled
with a boolean value $y_i$, that represents the supervision from the
user (i.e., $y_i=\top$ if $r_i$ and $r_i'$ are the same object,
$y_i=\bot$ otherwise). As usual when using distance-based
classification techniques, the underlying assumption is that the
embeddings of various instances of the same object are closer together
than those of different objects. Based on this assumption,
\textsf{getDecisionThresholds} sets the two thresholds so as to
maximize the probability that if $\delta$ is inside these boundaries,
the algorithm would take a wrong decision on the corresponding object,
given the information currently stored in memory. This is achieved by
solving the following optimization problem:

\begin{eqnarray}
\argmax_{\lambda_{l},\lambda_{u}}  &&  H(\calY_{\lambda_{l}}^{\lambda_{u}})  - H(\calY^{\lambda_{l}})  - H(\calY_{\lambda_{u}}) \label{eq:score} \\ 
\text{subject to: }  && \calY_{\lambda_{l}}^{\lambda_{u}} = \{ y | \langle y,\delta \rangle \in \calK \land \lambda_{l} \le \delta \le \lambda_{u} \} \nonumber \\
 		 && \calY^{\lambda_{l}} = \{ y | \langle y,\delta \rangle \in \calK \land \delta \le \lambda_{l} \} \nonumber \\
		  && \calY_{\lambda_{u}} = \{ y | \langle y,\delta \rangle \in \calK \land \lambda_{u} \le \delta  \} \nonumber \\
	&& \sum_{i=1}^{|\calK|} \mathbbm{1}(\lambda_{l} \le \delta_i \le \lambda_{u}) =  \ceil{\alpha |\calK|}\label{eq:effort}
\end{eqnarray}

\noindent Here $\calY_{\lambda_{l}}^{\lambda_{u}}$ is the set of user
answers related to objects with a distance within the two thresholds,
while $\calY^{\lambda_{l}}$ and $\calY_{\lambda_{u}}$ are the sets
related to objects below $\lambda_{l}$ and above $\lambda_{u}$
respectively. The function $H$ returns the entropy of a given set.
Eq.~\ref{eq:effort} imposes the constraint that the user is queried
with probability $\alpha$, where the probability is estimated using
the currently available feedback.  
The objective function is chosen in order to set the two thresholds in the
area where the algorithm is maximally confused, adjusting the size of
the area to the effort the user is willing to provide.  During
training, the algorithm will receive supervision only for those
examples where $\lambda_{l} \le \delta_i \le \lambda_{u}$.  Thus only
these elements will be added to the supervision memory $\calK$,
increasing the fraction of elements in $\calK$ having a distance
between the two thresholds. This, combined to the fact that $\alpha$
remains constant during training, will eventually lead to selecting
two closer and closer values for $\lambda_{l}, \lambda_{u}$, reducing
the size of the area of confusion and thus the probability to request
supervision. An alternative approach could be leaving the algorithm
the freedom to select the optimal size for the confusion area (leaving
$\alpha$ as an upper limit). We tested this approach, but we found
that this led to instability during training, due to poor estimates of
the correct size of the confusion area.

Another advantage of formulating the constraints as in
Eq.~(\ref{eq:effort}) is the ability to solve the maximization problem
efficiently, provided that the content of $\calK$ is stored inside a
list that is kept sorted with respect to $\delta_i$. We refer to this
list as $sorted(\calK)$. Then the optimal solution is associated to
one of the contiguous sub-lists of $sorted(\calK)$ of length
$\ceil{\alpha |\calK|}$. Let $\langle \delta_j, e_j \rangle$ and
$\langle \delta_k, e_k \rangle$ be the first and last element of a
sub-list $S$, where $j,k$ refer to the positions of the elements in
the full list $sorted(\calK)$, so that
$k = j + \ceil{\alpha |\calK|}$. Setting $\lambda_{l}^S = \delta_j$
and $\lambda_{u}^S = \delta_k$ guarantees that the constraint in
Eq.~(\ref{eq:effort}) is satisfied. Our algorithm evaluates the
objective~(\ref{eq:score}) for each of the contiguous sub-lists (there
are at most $|\calK|$ of them) and returns the values
$\lambda_{l}=\lambda_{l}^{S^*}, \lambda_{u}=\lambda_{u}^{S^*}$
corresponding to the sub-list $S^*$ for which the objective is
maximized. 

For evaluation purposes as well as to allow the algorithm to work in
absence of user supervision, it is useful to have a modality where
asking the user is not an option, and the algorithm can only decide
between recognizing the object as already seen or considering it a new
object. This can be done defining a single ``recognition'' threshold
$\lambda_r$, such that if $\delta$ is lower than the threshold the
object is recognized as the same as its nearest neighbor, otherwise it
is considered a new object. The threshold can be set in order to
maximize the number of correct predictions given the available
supervision $\calK$, by solving the following optimization problem:
\begin{equation}
  \label{eq:lambda_r}
  \lambda_r = \argmax_{\lambda}  \sum_{i=1}^{|\calK|} \mathbbm{1}(  (\delta_i < \lambda) \oplus \neg y_i )  
\end{equation}
\noindent where $\mathbbm{1}$ is the indicator function mapping true
to 1 and false to 0, and $\oplus$ is the exclusive OR. This problem
too can be solved in time linear in the size of $\calK$, by just
testing all thresholds
$\lambda^i = \frac{\delta_i + \delta_{i + 1}}{2}$ for
$i \in [0,|\calK|]$ (where $\delta_0=\delta_1 - \epsilon$ and
$\delta_{|\calK|+1}=\delta_{|\calK|}+\epsilon$ for an arbitrary small
  $\epsilon$).

\begin{figure*}[t]
\centering
  \begin{tabular}{cccc}
      \quad &\subfloat{\includegraphics[width = 0.3\textwidth]{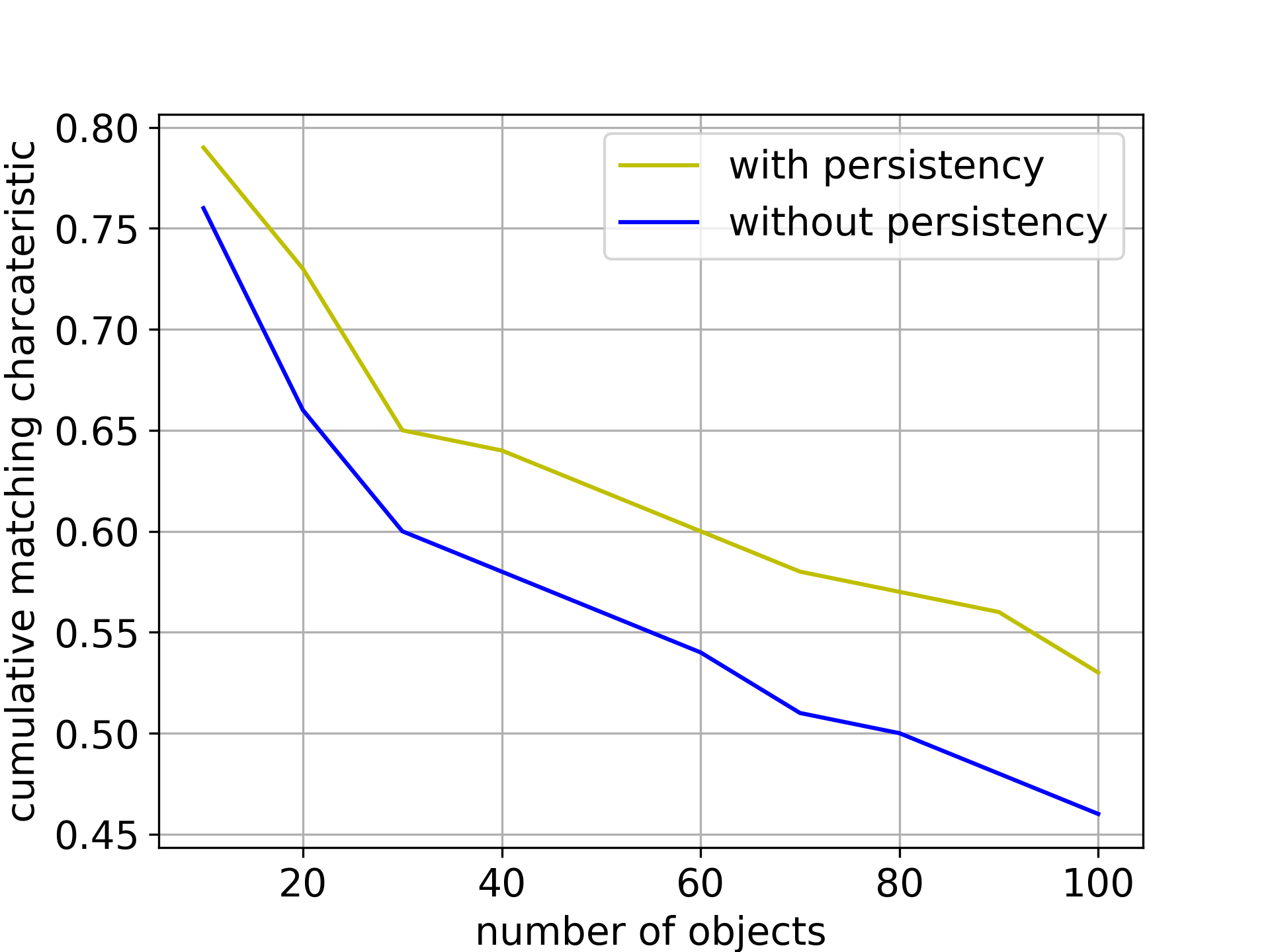}} & 
	  \subfloat{\raisebox{.07\height}{\includegraphics[width = 0.2\textwidth]{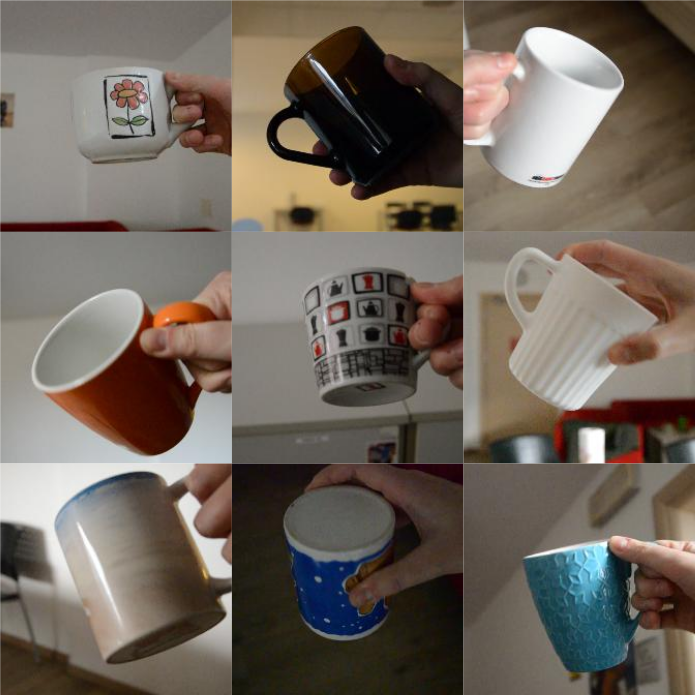}}} & 
	  \subfloat{\includegraphics[width = 0.3\textwidth]{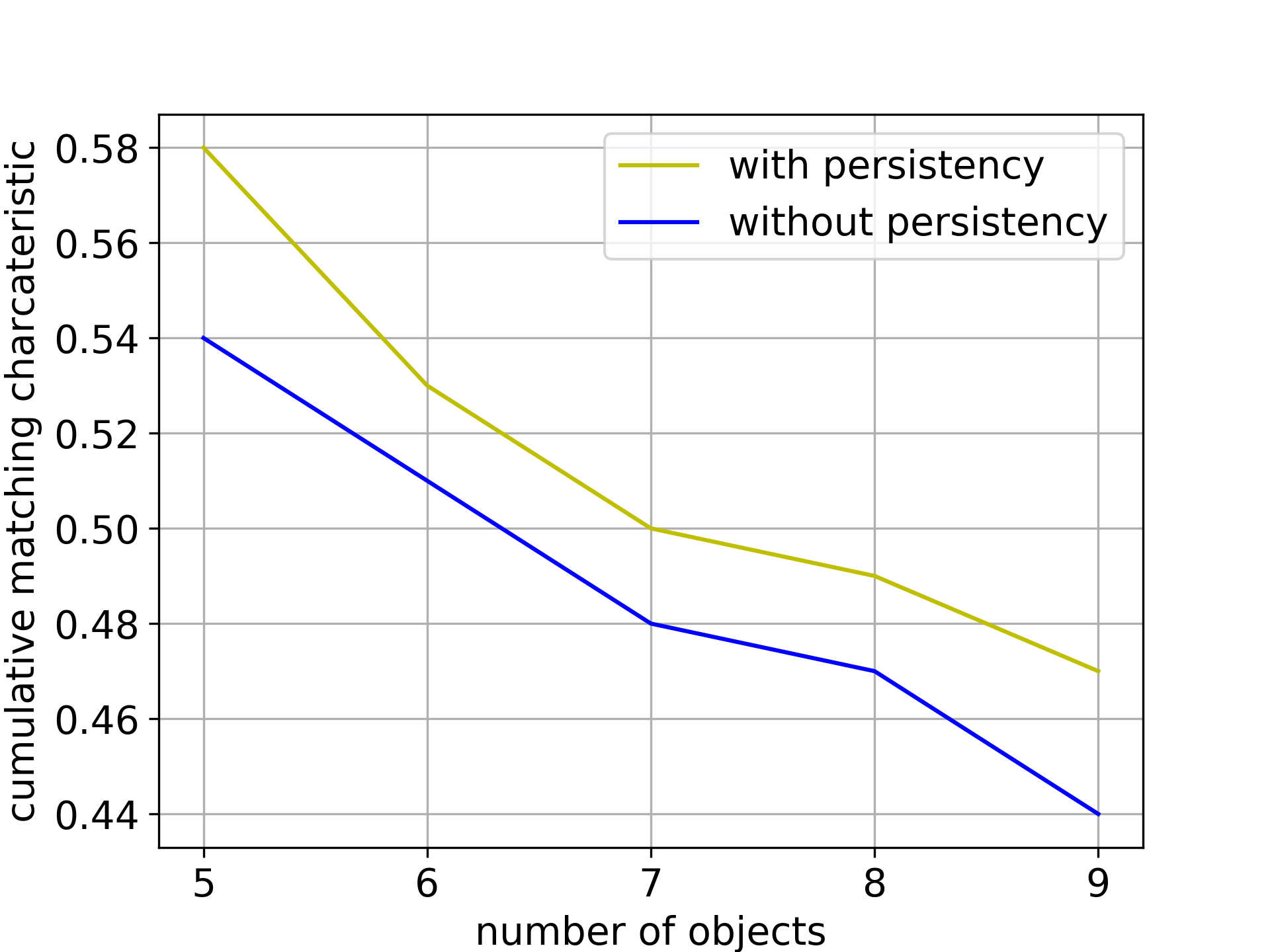}} \\ 
  \end{tabular}
  \caption{Re-identification results on generic objects (left) and on the coffee mugs (right) depicted in the middle.\label{fig:reid}}	
\end{figure*}
\section{Evaluation}

The framework described in Section~\ref{sec:framework} differs in many
aspects from the ones of mainstream classification algorithms. For
this reason, commonly used benchmarks meant for classification, even
those employed for continuous lifelong learning, aren't suited to
evaluate our approach. The ideal dataset should contain a collection
of videos, each focusing on  a single instance of an object, with the same
object appearing in multiple videos over different backgrounds,
somehow emulating the first stages of developmental learning under an
egocentric point-of-view~\cite{SmiSlo2017}. The egocentric perspective
was selected because it is the most natural setting for this task.  It
mimics the behavior of a user who focuses on an object she is
interested in, rotates and deforms different perspectives.

As far as we know only
two public datasets satisfy these requirements
\cite{ren2009egocentric,lomoncaco2017core50}.  The larger of the two,
called CoRe50 \cite{lomoncaco2017core50} contains a total of 50
objects, each recorded 11 different times.  Given that our goal is to
measure the ability to deal with the unknown in an incremental
scenario, we need to prioritize the number of different objects over
the number of recorded sequences.  Thus, we collected a new dataset of
videos of objects commonly found in office/home spaces. The dataset
contains 500 short videos of 100 different objects, 5 videos per
object, recorded with different backgrounds, and it is freely
available,
as well as the python code used to run the 
experiments \footnote{Code and dataset are available at: https://git.io/JvWUh}.
The main findings in this paper are however confirmed
when evaluating our algorithm on the CoRe50 dataset.

\subsection{The Role of Persistence}
Our first experiment is aimed at investigating how exploiting
persistence in space-time affects the recognition performance of our
model. We compare our algorithm with an alternative model that does
not make use of this feature.  Without persistence, each frame of a
sequence should be classified independently of the others, as if they
were a collection of pictures.  We performed these tests on a
closed-world re-identification setting, similar to the one used in
person re-identification \cite{li2018harmonious,wu2018exploit}.  The
aim of these tests is to evaluate the role of persistence
independently of the other aspects of our framework, like open-world
recognition and online active learning.

For each object in our dataset, we sampled one video for the training
set and one for the test set. For each test sample, we then computed
its nearest neighbor in the training set, where samples are videos
when using persistence, and individual frames otherwise.
Figure~\ref{fig:reid}(left) reports the fraction of videos (or frames)
that have as nearest neighbor a sample of the same object (i.e., we
report the Cumulative Matching Characteristic used in
re-identification, with a number of positions $k=1$), averaged over
100 folds, for an increasing number of objects to re-identify.  The
advantage of using persistence is apparent, and while increasing the
number of objects clearly decreases the performance of both
alternatives, the gap is substantially preserved.

\subsection{Instance-level Recognition}

The network we use for embedding frames (ResNet152) was originally
trained to classify objects of 1000 different classes of ImageNet, a
database that follows the same structure of the WordNet lexical
database.  Even if the original training set covered a broad spectrum
of the objects commonly found in tasks recognition (including some in
our own dataset), ResNet was trained with class-level supervision, a
way that explicitly pushes the network towards suppressing intra-class
variance.

In order to assess the performance of our algorithm at
instance-level object recognition, we performed a second series of
re-identification experiments focused on the ``coffee mug''
synset. Note that the synset appears as a leaf in the tree of
ImageNet, i.e., the network we use for the embedding is not trained to
distinguish between different mugs, but to consider them as a single
category. We recorder a small collection of videos of nine different
coffee mugs (shown in the middle of
Figure~\ref{fig:reid}). Figure~\ref{fig:reid}(right) presents the
re-identification results.
As expected, performance is lower than the ones for generic
objects, but the persistence model still obtains consistently better
results than the frame-based one.

\begin{figure*}[t]
	  \centering
	  \setlength{\tabcolsep}{0pt}
	\begin{tabular}{cccccc}	

		\rotatebox[origin=l]{90}{ \qquad\quad\random}  &$\;\;$ &
		\subfloat{\includegraphics[width = 0.3\textwidth]{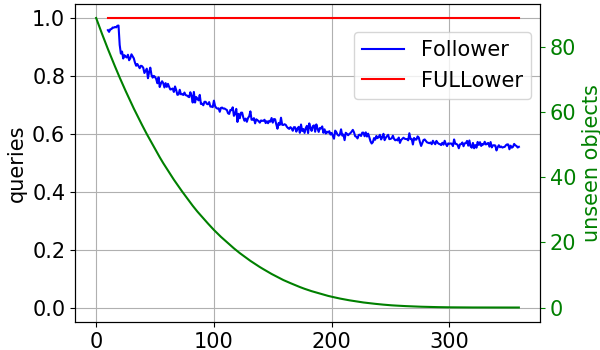}} & $\;\;$ &
      \subfloat{\includegraphics[width = 0.3\textwidth]{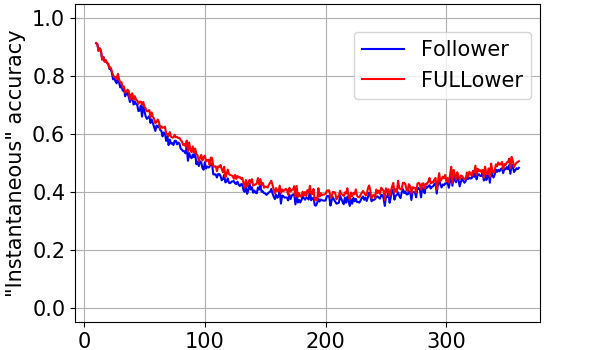}} &
		\subfloat{\includegraphics[width = 0.3\textwidth]{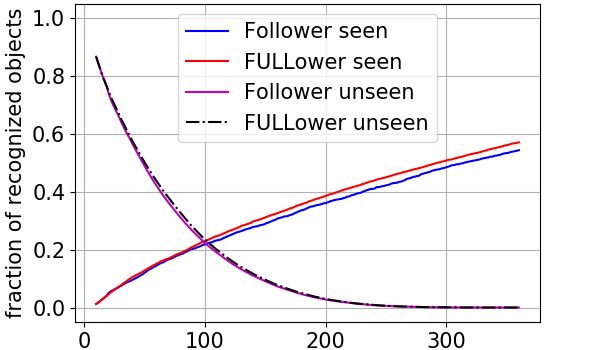}} 
		\\
		\rotatebox[origin=l]{90}{\qquad\qquad\devel}  &$\;\;$ &
		\subfloat{\includegraphics[width = 0.3\textwidth]{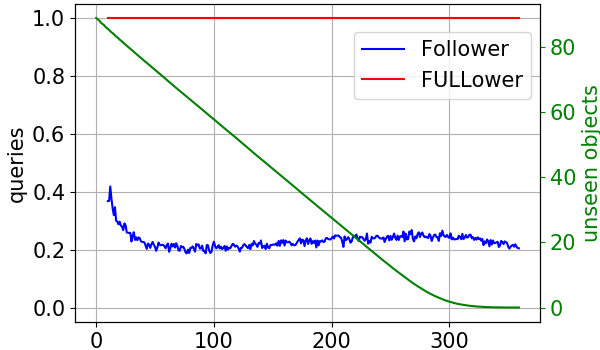}} &  &
      \subfloat{\includegraphics[width = 0.3\textwidth]{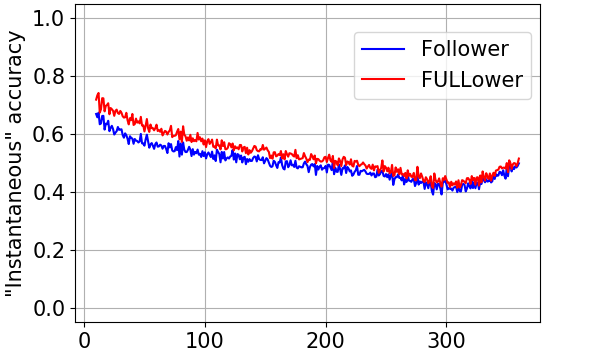}} &
	  \subfloat{\includegraphics[width = 0.3\textwidth]{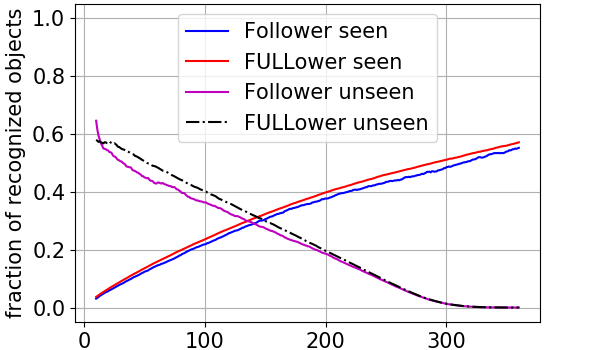}} 
  \end{tabular}
  \caption{Open world recognition results for increasing number of
    iterations.  \label{fig:openworld_res}}
\end{figure*}

\subsection{Open World Experiments}

The next series of experiment aims at evaluating the performance of
our algorithm in the setting described in Section
\ref{subsec:setting}.  One of the key elements that distinguish our
setting from a standard object recognition one is the fact that data
arrive as a continuous stream, as typical of online learning settings,
and the order in which data are presented to the algorithm can have an
impact on the quality of learning~\cite{Bengio:2009}. In infants, the
ratio at which novel objects appear is considered crucial for the
development of their recognition
capabilities~\cite{SmiSlo2017,Clerkin2017}. In order to investigate
how this can affect our recognition algorithm, we considered two
different policies for introducing novel objects. The fist one, named
\random, presents videos in a random order. The second, that we named
\devel (short for developmental), at each iteration picks the video of
a randomly chosen unseen object with probability 0.3, and randomly
selects the video of an already seen object otherwise.  We chose the
probability 0.3 to evenly spread the encounter of new objects.
Lowering the probability would result in showing all the videos of
already seen objects before introducing a new one. Raising the
probability would eventually result in showing all the unseen objects
before showing a video of an already seen object, resulting in a
behavior similar to the one obtained showing the sequences in random
order (or even more pronounced).  While being less intuitive, the
\devel setting is less artificial than the \random setting, especially
when the number of objects increases.  Assuming that the dataset
contains each object in the world, the \random setting would imply that
the user sees the majority of objects once, before interacting with an
already seen object.  That is not how a human interacts with the
environment and learns. In this sense, the \devel setting is clearly
a better approximation of the behavior of a human.

For each of the
two policies, we compared our recognition algorithm, \method (see
Algorithm~\ref{alg:incremental}) with a fully supervised version of
the algorithm (which we refer to as \fullmethod), where user feedback
is requested at each iteration.

The dataset was divided in three subsets.  First, 10 objects were
randomly selected, and all their sequences ($50$ in total) were used
to perform unsupervised evaluation after the online training phase.
For each of the remaining 90 objects, four videos were used for the
interaction with the algorithm, while the fifth was kept in a separate
set for online evaluation in the training phase. The procedure was
repeated 2000 times with different permutations of the data and
results were averaged.

\subsubsection{Supervised phase evaluation}
As \method needs as input the expected user effort $\alpha$, we
tuned it so as to ask the minimum supervision required to have similar
performance as \fullmethod. For the \random setting, this was achieved
with $\alpha = 0.92$, while in the \devel setting, $\alpha = 0.35$ was
sufficient. For the first 10 iterations
\method was forced to always ask for supervision, in order to 
bootstrap the estimation of its internal thresholds.

Figure~\ref{fig:openworld_res} shows the results of our experiments in
terms of user effort and recognition quality for an increasing number
of objects presented to the algorithms, for the \random (top row) and
the \devel (bottom row) setting respectively. The left column shows
the average fraction of objects never shown to the algorithm up to that
iteration included (green curve).
The plot also shows the number of queries to the
user for \method (blue curve) and \fullmethod (constantly one, red
curve); these  can be interpreted as the probability to 
request supervision at each iteration.
In the \random setting, in the first iterations the vast
majority of the videos contains brand new objects, while in the last
iterations the probability of encountering something new is close to
zero. The \devel setting spreads the encounter of objects evenly,
except for the last iterations when almost all objects have already
been seen. This latter setting is highly beneficial in terms of user
effort, as the amount of supervision that \method requests to match
the performances of \fullmethod is far less than the one it needs in
the random case, in agreement with what believed about infant
learning~\cite{Clerkin2017}.

The middle column in Figure~\ref{fig:openworld_res} shows the
``instantaneous'' accuracy in classifying the next object (as brand
new or as one already seen). This can be computed as:
\begin{equation}
\label{eq:instacc}
\mathbbm{1}\left(
\begin{gathered}
\delta(r_s,r_{min}) \le \lambda_r \land \same(r_s,r_{min})) \\
\lor \\
\delta(r_s,r_{min}) > \lambda_r \land \nexists \, r'_s \in \calM : \same(r_s,r'_s)
\end{gathered}\right)
\end{equation}
where $\lambda_r$ is the recognition threshold in
eq.~\ref{eq:lambda_r} (asking to the user is not an option here),
$\calM$ is the memory of the algorithm (at each training iteration),
$r_{min}$ is the nearest neighbor in $\calM$, and \same is true if
$r_s$ and $r_{min}$ are representations of the same object. For both
settings, the drop in performance follows the shape of the
probability of encountering new objects (that is, it decreases as the
number of objects to recognize increase, like in the re-identification
experiments shown in Fig.~\ref{fig:reid}). By design (i.e., choice
of the alpha values) the performance of the \methods never fall much
below the ones of the \fullmethods.  In both settings the performance
increase towards the end, due to the fact that the algorithms are
exposed mainly to objects they have already seen. 

We also evaluated the performance of the algorithms on a separate
\emph{in-training} evaluation set, as customary in online learning settings. 
The right column in Figure~\ref{fig:openworld_res} shows the fraction
of correctly recognized objects (in red for \fullmethod, in blue for
\method):
$$
\frac{1}{|\testset|} \sum_{s \in \testset} \mathbbm{1}\left(\delta(r_s,r_{min}) \le \lambda_r \land \same(r_s,r_{min})\right)
$$
and the fraction of examples correctly identified as unseen objects
(in black for \fullmethod, in purple for \method):
$$
\frac{1}{|\testset|} \sum_{s \in \testset} \mathbbm{1}\left(\delta(r_s,r_{min}) > \lambda_r \land \nexists \, r'_s \in \calM : \same(r_s,r'_s)\right)
$$
over the 90 hold-out videos $\testset$ (one per object). In the
\random setting, due to the fact that in the first iterations the
overwhelming majority of the encountered object were never seen
before, the two algorithms are strongly biased towards marking every
object as unseen. For this reason the black dot-dashed and purple curves are
overlapping. Thanks to the large amount of supervision, the recognition
performances over the already seen objects of \method never falls
too much below the one of \fullmethod.

In in \devel scenario the
algorithms are exposed to both new and unseen object right from the
beginning.  As \method requests supervision only for ambiguous cases
(i.e., nearest-neighbor distance neither low nor high enough), it is
more biased towards predicting objects as seen with respect to
\fullmethod. This, coupled with the fact that in this setting \method
requests supervision less than 30\% of the times, results in 
a slightly lower performance in recognizing unseen objects with respect 
to \fullmethod. As the number of iterations  increase, the performance
gap shrinks. For the same reason \method  
is able to closely match the performances of \fullmethod
in in recognizing already seen objects.

\begin{figure*}[h]
	  \centering
	  \setlength{\tabcolsep}{0pt}
	\begin{tabular}{cccccc}	

		\rotatebox[origin=l]{90}{ \qquad\quad\random}  &$\;\;$ &
		\subfloat{\includegraphics[width = 0.3\textwidth]{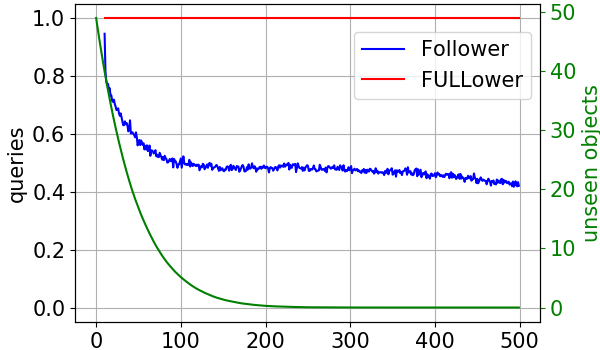}} & $\;\;$ &
      \subfloat{\includegraphics[width = 0.3\textwidth]{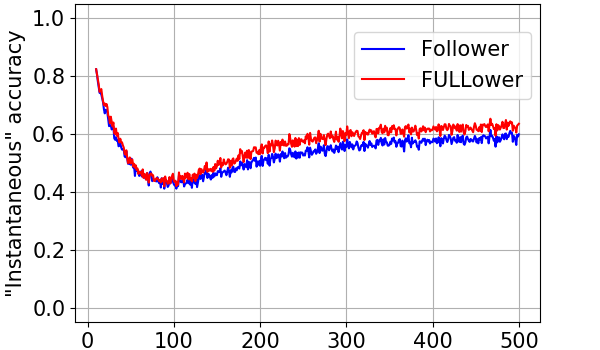}} &
		\subfloat{\includegraphics[width = 0.3\textwidth]{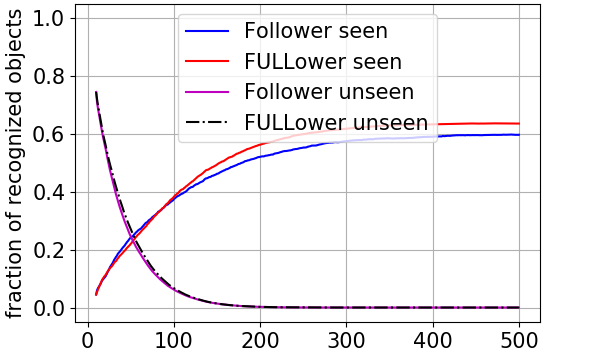}} 
		\\
		\rotatebox[origin=l]{90}{ \qquad\qquad\devel}  &$\;\;$ &
		\subfloat{\includegraphics[width = 0.3\textwidth]{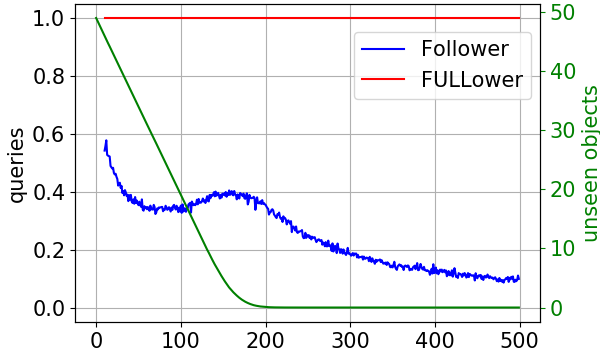}} &  &
      \subfloat{\includegraphics[width = 0.3\textwidth]{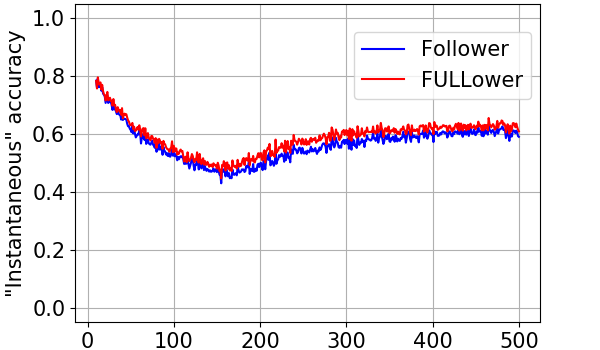}} &
	  \subfloat{\includegraphics[width = 0.3\textwidth]{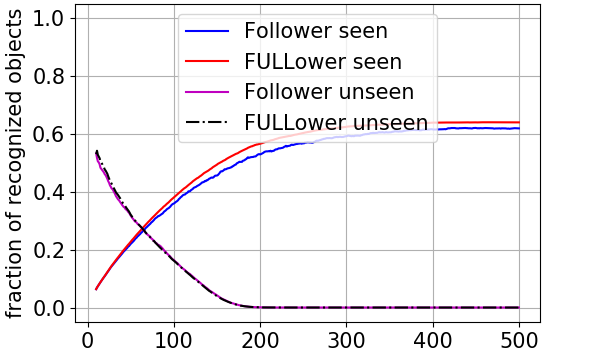}} 
  \end{tabular}
  \caption{Open world recognition results on the CoRe50 dataset for
    increasing number of iterations.  \label{fig:openworld_res_CoRe50}}
\end{figure*}

The same trends were found when repeating these experiments over the
CoRe50 dataset \cite{lomoncaco2017core50}.
Figure \ref{fig:openworld_res_CoRe50} presents the results
obtained over the \random and \devel setting, obtained using a value
of $\alpha$ of $0.84$ and $0.6$ respectively. These two values were
selected in order to minimize the amount of supervision in the two
setting while keeping the performances at a level comparable with
\fullmethod.

The results are in line with the ones obtained over our dataset. A \devel
setting still requires less effort from the user to enable \method to 
match the performances of \fullmethod. 
Due to the fact that CoRe50 contains many sequences but fewer different 
objects (compared to our dataset), 
in the $90\%$ of the iterations the models receive a sequence
containing an already seen object. For this reason the \method models
make less queries on average for each iteration in the experiments involving
this dataset.

The different number of sequences per object in the CoRe50 dataset,
with respect to our dataset, led to have a distribution of unseen
objects (the green curves in the first row of graphs) that is quite
similar between \random and \devel. In both settings almost all the
objects are shown to the models in the first 200 iterations. This is
due to our choice to keep the parameters of the \devel setting,
i.e. the probability of encountering a new object, at the same value
of the ones used in the experiments showed in the main paper.  As a
result, the graphs of the instantaneous accuracies and the recognition
performances of seen and unseen objects, comparing the two settings,
have a similar shape.

\subsubsection{Unsupervised phase evaluation}
The last series of tests are aimed at measuring the ability of the
model to autonomously recognize and discriminate new objects. These
tests are performed {\em after} the interactive training session with
the user. The model is presented with all the $50$ sequences of the
$10$ objects that were kept separate and never shown. The role of the
model is to classify and store them as in the interactive phase, but
without resorting on the supervision of the user. As for the training
phase, we tested the performance of our algorithm by showing these
$50$ sequences both with the \random and the \devel policy\footnote{Given 
the low number of different objects, we
decided not to compute the unsupervised evaluation 
on the CoRe50 dataset, in order to
avoid to further reduce the number of objects in supervised
evaluation}.

In this phase we computed the overall accuracy of the model, averaging
the instantaneous accuracy (eq.~\ref{eq:instacc}) over all $50$
decisions, one for each sequence. We refer to this metric as the
Averaged Instantaneous Accuracy (AIA).  As in this phase no
supervision is provided by the user, this evaluation can be seen as a
sort of incremental clustering. At each iteration the model observes a
new sequence and decides whether to consider the object in it as the
same as one seen previously (linking it to its nearest-neighbor in
memory), or to add it as a novel object. Let's consider a graph with
sequences as nodes and an edge between each pair of nodes predicted as
representing the same object. The connected components of this graph
can be seen as clusters, each representing a (presumably) distinct
object.  We extracted these clusters for the $50$ evaluation sequences
and we computed two widely used clustering metrics, the Adjusted
Mutual Information score (AMI) and Adjusted Rand Index score (ARI).

Table~\ref{tab:results}(top) shows the results obtained when
presenting evaluation sequences using the \random policy. The table
shows the metrics described above, computed over \method trained using
the \random (left) or the \devel (right) policy. In order to compare the
two training policies over the same number of examples, we tried
different values of $\alpha$ so as to end up with an approximately
equal number of training examples ($|\calK|$ in the table), for a
decreasing number of training examples. It is easy to see that when
trained with the \random policy, the performance of \method are
seriously affected by a decrease in the number of training
instances. On the other hand, the
\devel policy during training enables
the model to retain similar levels of accuracy even for substantially
reduced amount of user feedback. While the clustering performance are
more affected by the reduced level of supervision, this effect is much
less pronounced when using the \devel policy during training with
respect to the \random one.

Table~\ref{tab:results}(bottom) shows the results obtained when
presenting evaluation sequences using the \devel policy. The trends seen
with the \random evaluation policy are basically preserved here, with
the \devel policy at training time leading to better results than a
\random one. Comparing the two tables, it is apparent that \method
achieves massive improvements in recognition accuracy when evaluated
with sequences in the \devel rather than \random order, with up to 100\%
increase in AIA. Clustering performance are also substantially better
in most cases, especially when the model is trained with the \devel
policy. Note that the decrease in clustering performance when reducing
the number of training examples is more pronounced here with respect
to the top part of the table. This is due to the fact that the \devel
evaluation policy enables the method to substantially increase its
ability to recognize already seen objects, at the cost of a relative
decrease in ability to identify the unknown. In absolute terms,
however, also in terms of clustering the best results are achieved
when \method is both trained and evaluated with the \devel policy.

\begin{table}
  \caption{Test phase evaluation results. Top results refer to a
    \random evaluation policy, bottom ones to the \devel one. Results
    are computed over \method trained using  the \random (left) or a 
    \devel (right) policy.
  }
    \centering
\begin{tabular}{lcccc|cccc}
  & \multicolumn{4}{c}{\random} &\multicolumn{4}{c}{\dev} \\
  & $|\calK|$  & AIA & ARI & AMI & $|\calK|$  & AIA & ARI & AMI \\
  \parbox[l]{0mm}{\multirow{6}{*}{\rotatebox[origin=c]{90}{\random}}} 
  & 285 & 0.47 & 0.45 & 0.38 & 289 & 0.46 & 0.44 & 0.38 \\
	& 237 & 0.45 & 0.43 & 0.37 & 240 & 0.46 & 0.42 & 0.36 \\
	& 210 & 0.43 & 0.39 & 0.34 & 203 & 0.45 & 0.42 & 0.36 \\
	& 177 & 0.4 & 0.35 & 0.31 & 152 & 0.45 & 0.41 & 0.35 \\
	& 81 & 0.32 & 0.21 & 0.2 & 84 & 0.45 & 0.4 & 0.38 \\
	& 36 & 0.33 & 0.22 & 0.21 & 37 & 0.45 & 0.35 & 0.37 \\
        \hline 
  \parbox[l]{0mm}{\multirow{6}{*}{\rotatebox[origin=c]{90}{\dev}}} 
  & 285 & 0.73 & 0.56 & 0.5 & 289 & 0.73 & 0.6 & 0.54 \\
  & 237 & 0.73 & 0.49 & 0.45 & 244 & 0.73 & 0.56 & 0.52 \\
  & 210 & 0.72 & 0.44 & 0.4 & 203 & 0.73 & 0.55 & 0.5 \\
  & 177 & 0.71 & 0.37 & 0.35 & 152 & 0.73 & 0.52 & 0.49 \\
  & 81 & 0.68 & 0.21 & 0.2 & 84 & 0.73 & 0.44 & 0.45 \\
  & 36 & 0.69 & 0.22 & 0.21 & 37 & 0.72 & 0.35 & 0.4 \\  
\end{tabular}
\label{tab:results}
\end{table}

\section{Conclusion and Future Work}
\label{sec:conclusion}

In this paper we presented the first results of a long term project
whose goal is to build systems which see, perceive and interact in
open world environments like humans.
Our results show that \method is capable of progressively memorizing
novel objects, even in complete absence of supervision, and that a
developmental strategy is highly beneficial in boosting its
performance and reducing its need for human supervision.

In this line of thinking we see as strategic the possibility to make
this work more semantics and knowledge-aware, this being the basis for
a meaningful interaction with humans. The key choices underlying this
work (i.e., the exploitation of persistence, the choice of a
similarity-based approach and the attention to incrementality) were
indeed motivated by this intuition. The use of similarity and the
focus on the instance-level allow for the adaptability to an open
world, with instance-level recognition naturally scaled to the
perception of classes. The choice of the right granularity can be
adapted in an object-dependent way in order to contain
the ever increasing memory footprint, which is the cost to pay 
if the algorithms must keep recognizing more and more new objects.
Another approach would be to let  the user decide what she is interested in,
keeping in memory only information on the objects she cares, dropping or merging
the uninteresting objects.
This latter way highlights a key aspect of the egocentric recognition,
a setting where the classification capabilities of the machine is tailored
to fit the needs of a single human.

\ack This research has received funding from the European Union's Horizon 2020
FET Proactive project “WeNet – The Internet of us”, grant agreement No 823783.

\bibliographystyle{ecai}
\bibliography{bibliography}

\end{document}

%% file: pm_macros.tex
\usepackage{xspace}






%% file: paper.bbl
\begin{thebibliography}{10}

\bibitem{bendale2015towards}
A.~Bendale and T.~E. Boult, `Towards open world recognition', in {\em CVPR},
  pp. 1893--1902, (2015).

\bibitem{bendale2016towards}
A.~Bendale and T.~E. Boult, `Towards open set deep networks', in {\em CVPR},
  pp. 1563--1572, (2016).

\bibitem{Bengio:2009}
Y.~Bengio, J.~Louradour, R.~Collobert, and J.~Weston, `Curriculum learning', in
  {\em ICML}, ICML '09, pp. 41--48, (2009).

\bibitem{Clerkin2017}
E.~Clerkin, E.~Hart, J.~Rehg, C.~Yu, and L.~Smith, `Real-world visual
  statistics and infants' first-learned object names', {\em RSTB}, {\bf 372},
  (2017).

\bibitem{deng2009imagenet}
J.~Deng, W.~Dong, R.~Socher, L.~Li, K.~Li, and L.~Fei-Fei, `Imagenet: A
  large-scale hierarchical image database', in {\em CVPR}, pp. 248--255. IEEE,
  (2009).

\bibitem{foldiak91}
P.~Földiák, `Learning invariance from transformation sequences', {\em Neural
  Computation}, {\bf 3}(2),  194--200, (1991).

\bibitem{gepperth2016bio}
A.~Gepperth and C.~Karaoguz, `A bio-inspired incremental learning architecture
  for applied perceptual problems', {\em Cognitive Computation}, {\bf 8}(5),
  924--934, (2016).

\bibitem{goodfellow2014empirical}
I.~J. Goodfellow, M.~Mirza, D.~Xiao, A.~Courville, and Y.~Bengio, `An empirical
  investigation of catastrophic forgetting in gradient-based neural networks',
  in {\em ICLR}, (2014).

\bibitem{he2016deep}
K.~He, X.~Zhang, S.~Ren, and J.~Sun, `Deep residual learning for image
  recognition', in {\em CVPR}, pp. 770--778, (2016).

\bibitem{kirstein2009vision}
S.~Kirstein, A.~Denecke, S.~Hasler, H.~Wersing, H.~Gross, and E.~K{\"o}rner, `A
  vision architecture for unconstrained and incremental learning of multiple
  categories', {\em Memetic Computing}, {\bf 1}(4),  291, (2009).

\bibitem{koch2015siamese}
G.~Koch, `Siamese neural networks for one-shot image recognition', in {\em ICML
  Deep Learning Workshop}, (2015).

\bibitem{lee2017unsupervised}
H.~Lee, J.~Huang, M.~Singh, and M.~Yang, `Unsupervised representation learning
  by sorting sequences', in {\em ICCV}, pp. 667--676. IEEE, (2017).

\bibitem{li2018harmonious}
W.~Li, X.~Zhu, and S.~Gong, `Harmonious attention network for person
  re-identification', in {\em CVPR}, pp. 2285--2294, (2018).

\bibitem{lomoncaco2017core50}
V.~Lomonaco and D.~Maltoni, `Core50: a new dataset and benchmark for continuous
  object recognition', in {\em CoRL}, volume~78, pp. 17--26. PMLR, (13--15 Nov
  2017).

\bibitem{mishra2018simple}
N.~Mishra, M.~Rohaninejad, X.~Chen, and P.~Abbeel, `A simple neural attentive
  meta-learner', in {\em ICLR}, (2018).

\bibitem{misra2016shuffle}
I.~Misra, C.~L. Zitnick, and M.~Hebert, `Shuffle and learn: unsupervised
  learning using temporal order verification', in {\em ECCV}, pp. 527--544.
  Springer, (2016).

\bibitem{mu2017classification}
Xin Mu, Kai~Ming Ting, and Zhi-Hua Zhou, `Classification under streaming
  emerging new classes: A solution using completely-random trees', {\em IEEE
  Transactions on Knowledge and Data Engineering}, {\bf 29}(8),  1605--1618,
  (2017).

\bibitem{nakamura2012learning}
T.~Nakamura, K.~Sugiura, T.~Nagai, N.~Iwahashi, T.~Toda, H.~Okada, and
  T.~Omori, `Learning novel objects for extended mobile manipulation', {\em
  Robotic Systems}, {\bf 66}(1-2),  187--204, (2012).

\bibitem{li_unsupervised_2008}
L.~Nuo and J.~J. DiCarlo, `Unsupervised natural experience rapidly alters
  invariant object representation in visual cortex', {\em Science}, {\bf
  321}(5895),  1502{\textendash}1507, (sep 2008).
\newblock {PMID:} 18787171.

\bibitem{oliveira20163d}
M.~Oliveira, L.~S. Lopes, G.~H. Lim, S.~H. Kasaei, A.~M. Tom{\'e}, and
  A.~Chauhan, `3d object perception and perceptual learning in the race
  project', {\em Robotics and Autonomous Systems}, {\bf 75},  614--626, (2016).

\bibitem{parisi2018continual}
G.~I. Parisi, R.~Kemker, J.~L. Part, C.~Kanan, and S.~Wermter, `Continual
  lifelong learning with neural networks: A review', {\em arXiv preprint
  arXiv:1802.07569}, (2018).

\bibitem{prest2012learning}
A.~Prest, C.~Leistner, J.~Civera, C.~Schmid, and V.~Ferrari, `Learning object
  class detectors from weakly annotated video', in {\em CVPR}, pp. 3282--3289.
  IEEE, (2012).

\bibitem{redmon2016you}
J.~Redmon, S.~Divvala, R.~Girshick, and A.~Farhadi, `You only look once:
  Unified, real-time object detection', in {\em CVPR}, pp. 779--788, (2016).

\bibitem{ren2009egocentric}
X.~Ren and M.~Philipose, `Egocentric recognition of handled objects: Benchmark
  and analysis', in {\em CVPR Workshop}, pp. 1--8. IEEE, (2009).

\bibitem{rudd2018extreme}
E.~M. Rudd, L.~P. Jain, W.~J. Scheirer, and T.~E. Boult, `The extreme value
  machine', {\em TPAMI}, {\bf 40}(3),  762--768, (2018).

\bibitem{russakovsky2015imagenet}
O.~Russakovsky, J.~Deng, H.~Su, J.~Krause, S.~Satheesh, S.~Ma, Z.~Huang,
  A.~Karpathy, A.~Khosla, M.~Bernstein, et~al., `Imagenet large scale visual
  recognition challenge', {\em IJCV}, {\bf 115}(3),  211--252, (2015).

\bibitem{santoro2016meta}
A.~Santoro, S.~Bartunov, M.~Botvinick, D.~Wierstra, and T.~Lillicrap,
  `Meta-learning with memory-augmented neural networks', in {\em ICML}, pp.
  1842--1850, (2016).

\bibitem{scheirer2013toward}
W.~J. Scheirer, A.~de~Rezende~Rocha, A.~Sapkota, and T.~E. Boult, `Toward open
  set recognition', {\em TPAMI}, {\bf 35}(7),  1757--1772, (2013).

\bibitem{shen2017learning}
Y.~Shen, T.~Xiao, H.~Li, S.~Yi, and X.~Wang, `Learning deep neural networks for
  vehicle re-id with visual-spatio-temporal path proposals', in {\em ICCV}, pp.
  1918--1927. IEEE, (2017).

\bibitem{skovcaj2011system}
D.~Sko{\v{c}}aj, M.~Kristan, A.~Vre{\v{c}}ko, M.~Mahni{\v{c}},
  M.~Jan{\'\i}{\v{c}}ek, G.~M. Kruijff, M.~Hanheide, N.~Hawes, T.~Keller,
  M.~Zillich, et~al., `A system for interactive learning in dialogue with a
  tutor', in {\em IROS}, pp. 3387--3394. IEEE, (2011).

\bibitem{SmiSlo2017}
L.~B. Smith and L.~K. Slone, `A developmental approach to machine learning?',
  {\em Frontiers in Psychology}, {\bf 8},  2124, (2017).

\bibitem{snell2017prototypical}
J.~Snell, K.~Swersky, and R.~Zemel, `Prototypical networks for few-shot
  learning', in {\em NIPS}, pp. 4080--4090, (2017).

\bibitem{triantafillou2017few}
E.~Triantafillou, R.~Zemel, and R.~Urtasun, `Few-shot learning through an
  information retrieval lens', in {\em NIPS}, pp. 2255--2265, (2017).

\bibitem{wersing2007online}
H.~Wersing, S.~Kirstein, M.~G{\"o}tting, H.~Brandl, M.~Dunn, I.~Mikhailova,
  C.~Goerick, J.~Steil, H.~Ritter, and E.~K{\"o}rner, `Online learning of
  objects in a biologically motivated visual architecture', {\em Neural
  Systems}, {\bf 17}(04),  219--230, (2007).

\bibitem{wu2018exploit}
Y.~Wu, Y.~Lin, X.~Dong, Y.~Yan, W.~Ouyang, and Y.~Yang, `Exploit the unknown
  gradually: One-shot video-based person re-identification by stepwise
  learning', in {\em CVPR}, pp. 5177--5186, (2018).

\bibitem{yoon2018lifelong}
J.~Yoon, E.~Yang, J.~Lee, and S.~J. Hwang, `Lifelong learning with dynamically
  expandable networks', in {\em ICLR}, (2018).

\bibitem{zenke2017continual}
F.~Zenke, B.~Poole, and S.~Ganguli, `Continual learning through synaptic
  intelligence', in {\em ICML}, pp. 3987--3995, (2017).

\end{thebibliography}
